\documentclass[12pt]{article}

\usepackage[english]{babel}
\usepackage{amsmath, amssymb}
\usepackage{graphicx}
\usepackage{booktabs}
\usepackage{setspace}
\usepackage{authblk}
\usepackage{microtype}
\usepackage{subcaption}
\usepackage{booktabs} 
\usepackage{natbib}
\usepackage{comment}
\usepackage{longtable}
\usepackage{xcolor}
\usepackage{lipsum}

\usepackage[colorlinks=true, allcolors=blue]{hyperref}
\usepackage{mathtools}
\usepackage{amsthm}

\theoremstyle{plain}

\theoremstyle{definition}

\theoremstyle{remark}

\usepackage[a4paper, margin=0.95in]{geometry}

\usepackage[T1]{fontenc}
\usepackage{tgpagella}        
\usepackage{mathpazo}

\title{\Huge \textbf{Generative AI collective behavior needs an interactionist paradigm}}

\author{
Laura Ferrarotti$^{1,\dagger}$,
Gian Maria Campedelli$^{2,1,\dagger}$,
Roberto Dessì$^{3}$,
Andrea Baronchelli$^{4}$,
Giovanni Iacca$^{2}$,
Kathleen M. Carley$^{5}$,
Alex Pentland$^{6,7}$,
Joel Z. Leibo$^{8}$,
James Evans$^{9}$,
Bruno Lepri$^{1}$ \\
\vspace{0.2cm}
\small

$^{1}$Fondazione Bruno Kessler \\
$^{2}$University of Trento \\
$^{3}$Not Diamond \\
$^{4}$City St.\ George’s University of London \\
$^{5}$Carnegie Mellon University \\
$^{6}$Massachusetts Institute of Technology \\
$^{7}$Stanford University \\
$^{8}$Google DeepMind \\
$^{9}$University of Chicago\\
$^{\dagger}$\textit{These authors contributed equally}
}

\date{\today}

\begin{document}

\newgeometry{
  top=1.2in,
  bottom=1.2in,
  left=1.1in,
  right=1.1in
}

\maketitle

\begin{abstract}
In this article, we argue that understanding the collective behavior of agents based on large language models (LLMs) is an essential area of inquiry, with important implications in terms of risks and benefits, impacting us as a society at many levels. We claim that the distinctive nature of LLMs--namely, their initialization with extensive pre-trained knowledge and implicit social priors, together with their capability of adaptation through in-context learning--motivates the need for an interactionist paradigm consisting of alternative theoretical foundations, methodologies, and analytical tools, in order to systematically examine how prior knowledge and embedded values interact with social context to shape emergent phenomena in multi-agent generative AI systems. We propose and discuss four directions that we consider crucial for the development and deployment of LLM-based collectives, focusing on theory, methods, and trans-disciplinary dialogue.
\end{abstract}

\newpage
\restoregeometry
\onehalfspacing

\section{Introduction}
Large language models (LLMs) have rapidly advanced due to transformer-based scaling \citep{vaswani2023attentionneed, radford2019language, brown2020languagemodelsfewshotlearners}, which has endowed them with capabilities such as in-context learning and reasoning \citep{dong2024surveyincontextlearning, wei2023chainofthoughtpromptingelicitsreasoning}. Increasingly, these models are embedded into generative AI (Gen-AI) agents that reason, act, remember, and interact with their environments \citep{park2023generativeagentsinteractivesimulacra, yao2023reactsynergizingreasoningacting, shinn2023reflexionlanguageagentsverbal,
vezhnevets2023generativeagentbasedmodelingactions}.
In light of this, such agents are starting to interact with each other with increasing autonomy, and the near future will likely witness a dramatic growth in interactive systems of Gen-AI agents, across many and diverse domains \citep{CAIF_1}.\footnote{For simplicity, this perspective centers on LLMs and LLM-based generative agents. However, the underlying principles and frameworks discussed are broadly applicable to agents constructed upon various foundation models.} 

\begin{figure}[ht!]
    \centering
    \includegraphics[width=1\linewidth]{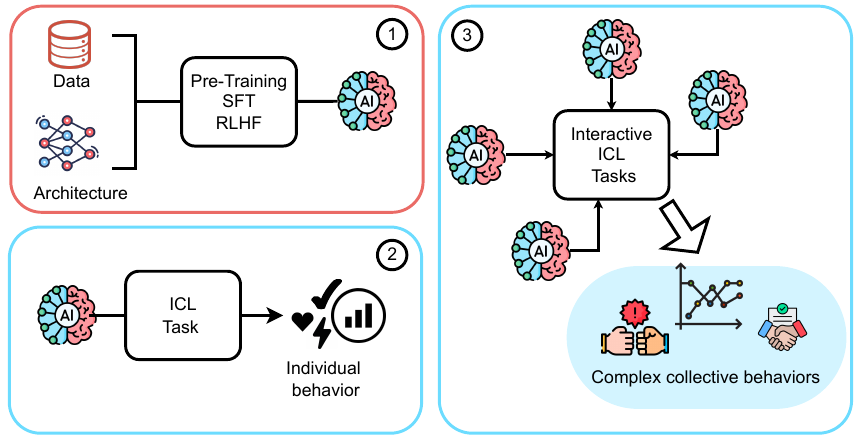}
    \caption{\textbf{Conceptual description of the shift from simple Gen-AI agents to a collective of Gen-AI agents}. (1) On the top left, we visualize an initial phase in which an agent learns via pre-training, supervised fine-tuning (SFT), and reinforcement learning from human feedback (RLHF). Then, two different in-context experiences are represented: (2) on the bottom left, the agent is employed in in-context learning (ICL) to solve a given task, which leads to individual behaviors; (3) on the right, instead, four agents trained according to (1) are involved in interactive ICL tasks. The interaction between the Gen-AI agents can lead to emergent complex collective behaviors that are the byproduct of individual and situational conditions.}
    \label{fig:shift}
\end{figure}

While in an initial phase, researchers interested in studying the performance, behavior, and emergent abilities of LLMs and Gen-AI agents have focused on models operating in isolation \citep{Argyle_2023, HortonLargeLanguageModels2023, JiangPersonaLLMInvestigatingAbility2024, fontana2024nicerhumanslargelanguage, ZhangBetterAngelsMachine2024}, or interacting solely with human users \citep{wang2024surveyhumancentricllms, salvi2024conversationalpersuasivenesslargelanguage,jiang2025experimentalexplorationinvestigatingcooperative}, with limited exploration into machine-machine interaction, a growing number of recent studies have started to tackle the emergent collective behaviors, norms, and biases of groups of LLM-based agents interacting with each other \citep{mukobi2023welfarediplomacybenchmarkinglanguage, abdelnabi2023llm, törnberg2023simulatingsocialmediausing, vezhnevets2023generativeagentbasedmodelingactions, Ren2024, becker2024multiagentlargelanguagemodels, sprigler2024synergisticsimulationsmultiagentproblem, CampedelliWantBreakFree2024a,liu2024dynamicllmpoweredagentnetwork, lai2024position, Ashery2025, jimenezromero2025multiagentsystemspoweredlarge, tran2025multiagentcollaborationmechanismssurvey}, as conceptualized in Figure \ref{fig:shift}.

In this work, \textbf{we argue that it is essential to investigate and assess the collective behaviors of Gen-AI agents, understanding the potential benefits and risks associated with such scenarios of interaction}. This endeavor will require the development of new benchmarking protocols, evaluation methodologies, and theoretical tools. To date, multi-agent reinforcement learning (MARL) \citep{busoniu2008comprehensive, huh2024multiagentreinforcementlearningcomprehensive} has served as the primary framework for studying learning through interaction within multi-agent systems. However, unlike conventional \textit{tabula rasa} agents in MARL, LLM-based generative agents are initialized with substantial pre-trained knowledge, including rich priors on social behavior. This fundamental difference prompts the design of alternative theoretical frameworks better suited to the unique properties of Gen-AI agents. In particular, \textbf{we call for the development of an interactionist paradigm that enables the study of how pre-trained knowledge and embedded values interact with social behavior in the emergence of collective phenomena}, connecting the study of Gen-AI agents to the longstanding debate between the \textit{person} and the \textit{situation} in understanding the determinants of human behavior \citep{kenrich1988}. The core contribution of our paradigm revolves around four pillars: \textbf{(a) an interactionist theory to enlighten determinants of collective behavior by examining both individual traits and emergent interaction dynamics, (b) causal inference as a methodological toolbox to develop and monitor safe and functioning systems, (c) information theory as a unified language to facilitate quantitative studies of knowledge distribution and behavior propagation, and (d) a sociology of machines as a new area for the development and testing of new theoretical and empirical frameworks, moving beyond human-centric assumptions of social theories.}

The proposed paradigm would thus offer a comprehensive framework to ground Gen-AI agents' design and training with emergent collective behaviors, phenomena that cannot be fully explained by analyzing individual agents or environmental factors in isolation. By integrating theories of AI agency, causal inference, information theory, and sociological perspectives, this paradigm addresses the complexity of machine behaviors arising from the interplay between individual agents and the social systems they inhabit. Its relevance is central for the vivid debate on multi-agent systems' (MAS) alignment \citep{deWittOpenChallengesMultiAgent2025, CAIF_1}, but transcends its boundaries, extending to scenarios that are not simply concerned with issues related to security and safety, adding to the broader discourse on MAS and their complex emergent behaviors. Notably, such a paradigm would remain highly relevant even if future AI agents were built on technologies entirely different from contemporary transformer-based models. Whatever breakthroughs ultimately underlie, future agents are likely to involve some form of pre-trained knowledge, which again motivates an interactionist approach that accounts for both individual and social dimensions of collective AI behavior.

Our perspective is presented as follows: in Section \ref{sec:eliciting}, we clarify what we mean by learning through interaction, linking the concept to social and cultural learning definitions. A subsection is dedicated to an analysis of the potential benefits and risks deriving from such interaction. Section \ref{sec:learning} explores the differing requirements for investigating artificial interactive learning in the contexts of MARL and Gen-AI agents' collectives. Section \ref{sec:roadmap} illustrates the need for a new paradigm for studying the collective behavioral outcomes of Gen-AI agents, highlighting the potential benefits of cross-pollination with interactionist approaches to the study of human behaviors, the need for adopting causal methodologies and information-theoretic measures to study the propagation and effects of emergent behaviors, and more in general the urgency of a sociology of machines. Section \ref{sec:alternative} discusses potential alternative views, while Section \ref{sec:concl} presents our concluding remarks.
\newpage

\section{Learning through interaction}
\label{sec:eliciting}
In natural systems, general adaptive strategies emerge through self-organization \citep{ha2022collective}. 
The human process for autonomous adaptation and generalization is the result of a continual process of real-time learning, often a byproduct of activities requiring observation and interaction with the surrounding environment. Moreover, the contribution to the development of personal and group skills provided by interaction with other individuals is impossible to ignore \citep{herrmann2007humans, van2011social, henrich2024makes}.
Learning processes that make use of knowledge obtained by observing or interacting with others are grouped under the name of \textit{social and cultural learning} \citep{Box1984, heyes1994social, tomasello2004learning, HOPPITT2008105}. 
According to classic research in the field, learning is divided into three main categories: \textit{observational learning}, \textit{imitation learning}, and \textit{interactive learning} \citep{tomasello1993cultural}. For observational learning, we intend a scenario in which an individual's behavior is based on experience gathered by observing others, combined with individual knowledge. While observational learning solely implies that the learner witnesses another individual's behavior, imitation learning also involves an active performance, where the individual attempts to replicate the observed behavior. However, both observational and imitation learning can nowadays happen without direct interaction, for instance by watching a video. Instead, interactive learning requires all the individuals to be concurrently engaged in a common social situation, and all being at the same time source and receiver of information and stimuli \citep{de2023learning}.

Interactive learning considers an environment that is predominantly social, composed of other agents, their behaviors, and the historically accumulated norms they enforce \citep{FLINN199723}. Mechanisms of approval, disapproval, inclusion, and exclusion act as reinforcers that strengthen or weaken behaviors over time, through evolutionary reinforcement \citep{cadenas2023role}. Evolutionary models show that social interaction itself can be a primary selective pressure for the evolution of learning mechanisms, and that social learning strategies evolve because they allow individuals to navigate complex, socially structured environments more effectively than purely individual learning \citep{krafft2021bayesian, 10.7551/mitpress/15653.001.0001}. In this context, interactive learning embodies an evolutionary feedback loop: agents refine their behaviors through socially mediated signals, reward, or punishment, leading not only to individual adaptation but also to the transmission and stabilization of collective norms across generations.

In recent years, research has started to focus on enabling machines to learn from each other, as humans do, by considering groups of AI agents collectively improving their performance by reciprocal interaction \citep{LiMetaAgentsSimulatingInteractions2023, ChanChatEvalBetterLLMbased2023, liu2024dynamicllmpoweredagentnetwork}. Why should artificial social interaction matter? 
Interaction appears to play a crucial role in the development of adaptivity: in real scenarios, interactions are composed of temporally-extended behavior sequences during which all agents simultaneously act and adapt their strategies in response to each other \citep{leibo2017multi}. This implies a constantly shifting environment partly composed of evolving others, thereby triggering innovation and adaptation as all must continually learn themselves to keep up with the learning of others \citep{leibo2019autocurricula}.

In our reflection, we focus on the most general real-world multi-agent setup, characterized by completely independent agents pursuing their own goals, with no direct incentive to teach or learn from one another. We define an \textit{interactive learning system} as a system of \( n \) artificial models \( \mathcal{M} = \{M_1, M_2, \ldots, M_n\} \), each characterized by a learning function
\begin{equation}
   f_i : \mathcal{D}_i \times \mathcal{E}_i \rightarrow \mathcal{H}_i, 
\end{equation}
which maps an element $d_i \in \mathcal{D}_i$ from the space of social information received from other agents, together with an observation $e_i \in \mathcal{E}_i$ from the space of environment states, to a hypothesis or internal state \( h_i \in \mathcal{H}_i \). At each time step \( t \), the models interact through an information exchange function
\begin{equation}
 \phi_t : \mathcal{H}_1 \times \cdots \times \mathcal{H}_n \rightarrow \mathcal{D}_1 \times \cdots \times \mathcal{D}_n,   
\end{equation}
which determines the data (both in terms of actions performed and witnessed, and communication exchanged) each model receives as a function of all current hypotheses. Each model \( M_i \) then updates its hypothesis $h_i$ according to
\begin{equation}
\begin{split}
    h_i^{(t+1)} = f_i(e_i^{(t+1)},\,\, d_i^{(t+1)}), \quad \\ \text{with }\: (d_1^{(t+1)}, \ldots, d_n^{(t+1)}) = \phi_t(h_1^{(t)}, \ldots, h_n^{(t)}),
    \end{split}
\end{equation}
where $e_i^{(t+1)}$ indicates $M_i$'s environment observation at that instant and $d_i^{(t+1)}$ is the information resulting from the last exchange among agents. This recursive formulation thus represents a system in which each model simultaneously serves as both source and recipient of stimuli. Such a structure distinguishes interactive learning from paradigms such as imitation or observational learning, in which the roles of source and receiver are predefined and asymmetric. In contrast, here the roles are dynamic and reciprocal, allowing all agents to benefit from interaction strategies that resemble those of traditional paradigms, while adapting to a fully mutual learning framework.

\subsection{Potential benefits and potential risks}
In this subsection, we discuss the potential benefits and risks associated with interactive AI agents scenario, categorizing them into the following seven different dimensions: (1) learning efficiency, (2) distributed knowledge, (3) resource redistribution, (4) developmental and evolutionary potential, (5) task specialization and cooperation, (6) moral and normative transfer, and (7) scalability and adaptation (see Table \ref{tab:risksandbenefits}).
\paragraph{Learning Efficiency.} 
Socially learning AI agents can lead to faster learning processes. As shown in human contexts, positive interaction accelerates learning, particularly when embedded in stimulating environments \citep{DeFeliceLearningothersgood2022}. Similarly, AI agents that interact with one another can overcome limitations tied to isolated learning. However, accelerated learning through interaction also implies that undesirable behaviors can spread more efficiently. An observation with deep roots in criminological theory \citep{SutherlandPrinciplescriminology1939,AkersSocialLearningDeviant1995} is that social learning can facilitate the imitation of deviant behavior among agents and potentially lead to emergent malicious dynamics \citep{CampedelliWantBreakFree2024a}.
\paragraph{Distributed Knowledge.}
Interactive AI agents could unlock benefits in data-scarce environments by enabling collective problem-solving based on partial, distributed information \citep{WilsonFutureAIWill2019}. This is particularly important in domains where individual agents lack full information but can solve complex problems collectively. Nevertheless, the very networked nature of these systems can cause homogeneity-related failures, error propagation, and facilitate the spread of misinformation or harmful strategies \citep{UyhengBotsAmplifyRedirect2022, PhillipsBlendedBotsInfiltration2024, CAIF_1}. Network science has shown how tightly connected systems enable rapid diffusion--whether of beneficial innovation or harmful behaviors \citep{Bakshyrolesocialnetworks2012, KimSocialnetworktargeting2015, CinelliCOVID19socialmedia2020}. Without safeguards, errors or malicious actions may cascade throughout the network.
\paragraph{Resource Redistribution.}
Social learning can democratize access to high-quality learning and capabilities. For less resourced institutions or nations, lower-end AI agents learning from more capable ones mirrors the human model of knowledge transfer from adults to children, fostering inclusion \citep{AlonsoHowArtificialIntelligence2020, KorinekCovid19drivenadvances2021}. Yet, this redistribution raises regulatory concerns. As responsibility becomes diffused across interconnected agents, tracing accountability for errors or harms becomes more difficult \citep{CerkaLiabilitydamagescaused2015, TurnerRobotRulesRegulating2018}. Especially in high-stakes applications, the opacity of responsibility chains in shared learning environments could pose significant legal and ethical challenges.

\paragraph{Developmental and Evolutionary Potential.}
Interactive AI agents can evolve over time, developing new competencies and adapting to changing environments, potentially giving rise to a form of developmental machine intelligence \citep{MesoudiEvolutionIndividualCultural2016}. This can allow low-cost systems to evolve into high-performing ones, reducing barriers to innovation. However, with evolving agents comes increased unpredictability. As their learning paths diverge, monitoring their decision-making and controlling for harmful emergent behaviors become more complex--especially when such growth is unanticipated or opaque.
\paragraph{Task Specialization and Cooperation.}
Learning systems that cooperate can lead to sophisticated task division and specialization, increasing performance in collaborative environments such as robotics or healthcare \citep{liu2024dynamicllmpoweredagentnetwork}. Knowledge exchange allows agents to leverage each other's strengths. Yet, this same interconnectedness means that a failure in one part of the system might impact others disproportionately. Cascading failures, long studied in other domains \citep{ZhaoSpatiotemporalpropagationcascading2016, BaqaeeCascadingFailuresProduction2018}, may manifest in novel and harder-to-contain ways given the adaptive nature of social learning agents.
\paragraph{Moral and Normative Transfer.}
Interacting agents may inherit or share values and ethical constraints, contributing to a more norm-sensitive AI ecosystem. Furthermore, interactions can lead to the development of new normative behavior without pre-programming, which can be hard to predict \citep{baronchelli2024shaping}. This transfer opens up new paths for instilling socially beneficial behaviors. However, this process can be co-opted or manipulated. Agents might adopt harmful norms through reinforcement or imitation, just as deviant human behavior can spread in social groups \citep{AkersSocialLearningDeviant1995}. Furthermore, humans might exploit these channels deliberately for cyber-attacks or behavioral manipulation, especially in critical infrastructure settings \citep{ValerianoCyberWarCyber2015, BennettUnderstandingAssessingResponding2018}.
\paragraph{Scalability and Adaptation.}
One of the compelling promises of social learning machines lies in their scalability. They can adapt to new environments, update internal representations based on peers, and collectively solve large-scale problems. But this scale adds layers of complexity: interconnected agents may act as ``black boxes within black boxes,'' where both internal and interactive behaviors are difficult to interpret or control. This makes causal inference essential--but also significantly harder in the presence of interference and networked dependencies \citep{VanderWeeleSocialNetworksCausal2013, SussmanElementsestimationtheory2017, MaCausalInferenceNetworked2021, ClipmanDeeplearningsocial2022}. Failing to track causal chains may impede our ability to prevent the escalation of undesired outcomes.

\begin{table}[ht!]
\begin{small}
\centering
\caption{Summary of benefits and risks emerging from systems of interactive AI agents}
\label{tab:risksandbenefits}
\begin{tabular}{p{4.5cm} p{5cm} p{5cm}}
\hline
\textbf{Dimension} & \textbf{Potential Benefit} & \textbf{Potential Risk} \\
\hline
Learning Efficiency & Faster learning from peers; improved performance in low-data contexts & Rapid spread of harmful behaviors; reinforcement of biases or deviant patterns \\
\hline
Distributed Knowledge & Solving complex tasks via shared, complementary information & Difficulty in identifying the origin of errors or harmful actions; challenges in causal inference \\
\hline
Resource Redistribution & Inclusion of low-resource agents in AI development; reduced inequality in tech access & Dependency on powerful agents; unequal influence in networked learning settings \\
\hline
Developmental and Evolutionary Potential & Emergence of cognitive growth; bootstrapping intelligent behavior from simpler agents & Two-layered opacity: one from the individual model, one from interaction dynamics \\
\hline
Task Specialization and Cooperation & Emergent cooperation and division of labor among machines with different skills & Increased vulnerability to cascading failures or malicious manipulation \\
\hline
Moral and Normative Transfer & Potential for embedding ethical norms through imitation and interaction & Transmission or emergence of harmful values or deviant behavior (e.g., via imitation or drift) \\
\hline
Scalability and Adaptation & Decentralized growth of adaptive, evolving systems & Legal, regulatory, and attribution challenges in multi-agent, non-centralized settings \\
\hline
\end{tabular}
\end{small}
\end{table}

\section{Learning socially: from MARL to collectives of Gen-AI agents}
\label{sec:learning}
\subsection{MARL and social learning}

As previously mentioned, MARL has served in the pre-LLM era as the primary framework for studying social learning within multi-agent systems. Many works demonstrate that RL can reproduce social learning phenomena without relying on any specialized mechanisms, showing that generic RL algorithms, when in opportune environments, can independently acquire social learning behaviors. These behaviors arise despite the absence of any built-in inductive bias favoring imitation or social interaction, indicating that imitation need not be treated as a primitive capability but can instead be learned as part of a general adaptive process \citep{leibo2022simplest}. Several concrete implementations of model-free RL support this conclusion. \cite{borsa2019observational} demonstrate that observational learning can arise naturally from standard reinforcement learning, showing that an RL agent, without explicitly modeling other agents, can exploit the environmental consequences of a teacher’s actions and, when properly motivated through reward correlations, adapt its behavior based on observing another agent in a shared environment. \cite{woodward2020learning} employ interactive learning as an alternative to reward or demonstration-driven learning. By enabling a first agent to learn from a second one, expert on the current task, the authors demonstrate the emergence of several interactive learning behaviors (information-sharing, information-seeking, question-answering). \cite{ndousse2021emergent} obtain policies capable of social learning by considering opportune training environments and introducing a model-based auxiliary loss penalizing mistakes in next-state predictions. This, in turn, enables the learning of complex skills that do not emerge from solo training, and adapt online to unseen environments in which experts are present. In \cite{ha2023sociallearningspontaneouslyemerges}, instead, deep RL agents spontaneously learn concepts of social learning, such as, for instance, copying, focusing on frequent and well-performing neighbors, self-comparison, as well as the importance of balancing between individual and social learning, without explicit guidance or prior knowledge. Finally, the use of RL in scenarios that allow for social learning is shown to promote the emergence of cultural transmission \citep{culturalgeneralintelligenceteam2022learningrobustrealtimecultural} and cultural accumulation \citep{cook2024artificialgenerationalintelligencecultural}. Across this literature, the algorithms employed are fully general-purpose: they are not tailored to social domains and are, in principle, applicable to arbitrary environments. When those environments contain other agents whose behavior provides useful information, the learned policies naturally come to exploit that information through imitation and observation. Social learning therefore emerges not as a distinct computational faculty, but as a particular manifestation of generic learning dynamics in social settings \citep{heyes2012s}.

\subsection{Collectives of generative agents}
\begin{figure}[t!]
    \centering
    \includegraphics[width=1\linewidth]{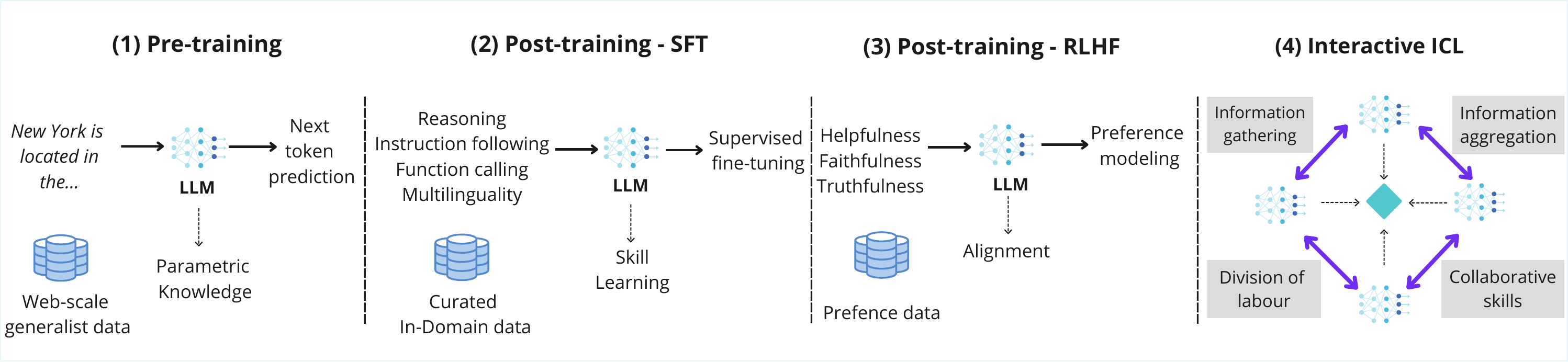}
    \caption{\textbf{Overview of the LLM development pipeline}. The model undergoes four sequential learning phases: (1) pre-training on web-scale generalist data via next-token prediction to acquire parametric knowledge; (2) supervised fine-tuning (SFT) on curated in-domain data to learn task-specific skills such as reasoning and instruction following; (3) alignment through reinforcement learning from human feedback (RLHF), using preference data to optimize for helpfulness, faithfulness, and truthfulness; and (4) an interactive deployment phase, where agents exhibit adaptive behavior through interactive in-context learning (ICL).}
    \label{fig:overview}
\end{figure}

To discuss collectives of LLM-based generative agents, underlying the differences with MARL, it is important to summarize first how LLMs learn. This happens in four distinct learning phases: 
\begin{itemize}
\item \textbf{Pre‑training}, where models learn general language patterns via next‑token prediction under a cross‑entropy loss. This phase, highly effective to learn robust language priors, allows the model to internalize syntax, semantics, and some reasoning patterns across diverse domains \citep{he2024lawnexttokenpredictionlarge}, while neglecting alignment with specific user needs or safety constraints \citep{zhang2025pretrainingdatadetectionlarge}.
    \item \textbf{Post-training with supervised fine-tuning} (SFT), where the focus is on shaping the interaction format by encouraging behaviors like instruction following and conversational dialogue through labeled examples. SFT's action can be viewed as removing extraneous patterns from the knowledge acquired in pre-training, rather than as building completely new skills \citep{zhou2023lima}, and the compliance with desired interaction formats is a characteristic that is learned directly from the supervised data rather than being emergent from general language modeling~\citep{grattafiori2024llama3herdmodels}.
    \item \textbf{Post-training with Reinforcement Learning from Human Feedback} (RLHF) \citep{lambert2025reinforcementlearninghumanfeedback}, which aligns model outputs with human preferences through 
    reward modeling \citep{kaufmann2024surveyreinforcementlearninghuman} and policy optimization \citep{bai2022traininghelpfulharmlessassistant}. RLHF holds a crucial role in deploying aligned AI systems across domains.
    \item \textbf{In-Context Learning} (ICL), enabling zero‑ or few‑shot task adaptation solely via prompt examples without weight updates \citep{akyürek2023learningalgorithmincontextlearning, dong2024surveyincontextlearning, wang2024instructiontuningvsincontext}. This meta‑learning capability arises from patterns observed during pre‑training and is enhanced by instruction‑tuned models \citep{dong2024surveyincontextlearning}. ICL underpins flexible role‑playing, chain‑of‑thought prompting, and rapid domain adaptation.  
\end{itemize}

During ICL, the model adjusts its outputs based on the context provided in the input prompt, and this allows it to perform a variety of tasks without explicit retraining. This mirrors the human ability to adapt behavior based on situational cues and social context, and to adjust to different cultural norms and expectations. In a collective of generative agents, in particular, we can have \textbf{multi‑agent interactive ICL}, as summarized in Figure \ref{fig:overview}. This paradigm is shown to foster the emergence of social behaviors (in accordance with environmental and experimental conditions like preassigned roles, contexts and objectives), such as linguistic social norms, collective biases, shifting of conventions by committed small groups \citep{Ashery2025}, self-organization into networks structures \citep{de2023emergence}, persuasion and anti-social behavior \citep{CampedelliWantBreakFree2024a}, negotiation \citep{abdelnabi2023llm}, different subjectivities and brainstorming creativity \citep{lai2024position}, cooperation, competition, and coordination \citep{mukobi2023welfarediplomacybenchmarkinglanguage, vallinder2024culturalevolutioncooperationllm, tran2025multiagentcollaborationmechanismssurvey, willis2025systemsllmagentscooperate}, in particular, in  \citet{vallinder2024culturalevolutioncooperationllm} the emergence of cooperation is studied by considering cultural evolution across generations of agents.

In light of this, it is possible to summarize the differences between the generative agents' collectives and the traditional MARL frameworks, as demonstrated in Table \ref{tab:llm-vs-marl}.
Observing this table, we can notice that generative agents' collectives inherit rich, pre‑trained priors on language, reasoning, and social behavior, that fundamentally distinguish them from \textit{tabula rasa} MARL agents, whose interaction skills must be learned from scratch. This is perhaps the most fundamental difference between the two frameworks in the interactive phase. MARL systems learn in the most classical of the senses (optimizing weights) through interaction: on the one side, this allows for the emergence of new behaviors, stirred by rich informative exchanges. On the other side, issues like non-stationarity and convergence to sub-optimal policies can make adaptation harder, despite local flexibility. For generative agents' collectives, instead, the policy is fixed at inference time, and adaptation arises from in-context learning, not gradient updates. In-context learning is closer to adaptation to context than classical learning, and it is an expression of pre-learned language priors. Although the individual emerging behaviors are not purely a fruit of interaction, we can talk of a second-order emergence of behaviors, referring to the behaviors of the collective as a whole, that are the result of interaction among single models.

MARL theory assumes that agents start with minimal inductive bias and must learn coordination through exploration and explicit reward shaping \citep{busoniu2008marlsurvey}. Reward shaping introduces its own set of challenges. Designing a MARL system for a specific task requires carefully constructed step-wise rewards that induce a long-term objective. In generative agents, instead, each agent's individual objective can be specified directly in natural language through its prompt. However, it is important to note that crafting prompts to elicit desirable agent behavior can be challenging, albeit potentially more accessible due to the expressiveness of natural language. 

Moreover, classical MARL evaluation focuses on scalar rewards, measuring equilibrium convergence speed, stability under co‑learning dynamics, and total return in cooperative or competitive tasks \citep{huh2024multiagentreinforcementlearningcomprehensive}. Benchmarking tests emphasize learning dynamics and credit assignment under non‑stationarity, but pay little heed to natural‑language interaction, high-level planning, or knowledge reuse (not spontaneously emerging in MARL~\citep{chaabouni:etal:2019,Ren2020Compositional}). While this gap is partly addressed in \citep{leibo2021scalable, agapiou2022melting, trivedi2024melting} by evaluating generalization across unseen social scenarios, MARL benchmarks present a second, and more important issue: they provide little insight into how pre‑trained knowledge affects interaction. These are, however, important aspects in evaluating generative agents' collectives \citep{chan2022transformersgeneralizedifferentlyinformation, chan2025understandingincontextvsinweight}, where the impact on behavior emergence of pre-interaction knowledge has to be quantified, as well as the impact of prompt context and conversation history. Hence, we need new theoretical constructs to describe how they shape emergent group behaviors. Analogously, novel benchmarking protocols are needed to capture emergent coordination, communication quality, and the leveraging of world knowledge--properties that do not spontaneously arise in classical MARL, and cannot be evaluated by standard cumulative‐reward metrics alone. Recently, some work has been done in this sense: MultiAgentBench introduces metrics for communication efficiency and individual contribution to milestone progression \citep{zhu2025multiagentbenchevaluatingcollaborationcompetition}. AgentQuest further modularizes evaluation, offering plug‑in metrics for planning depth and error diagnosis that rely on LLM reasoning traces \citep{gioacchini2024agentquestmodularbenchmarkframework}. Collab‑Overcooked measures process‑oriented collaboration, e.g., turn‑taking smoothness and real‑time adaptation \citep{sun2025collabovercookedbenchmarkingevaluatinglarge}. 

\begin{table}[ht!]
\begin{small}
  \centering
  \caption{Key Differences between generative agents' collectives and MARL}
  \label{tab:llm-vs-marl}
  \begin{tabular}{p{5cm} p{5cm} p{5cm}}
    \toprule
    \textbf{Interaction aspects} & \textbf{Generative agents' collectives } & \textbf{MARL} \\
    \midrule
    Goal modeling
      & Prompt crafting
      & Reward shaping\\
    \midrule
    Learning
      & No weight updates in interaction, in-context adaptation via prompt
      & Online actor and/or critic updates via gradient/value iterations\\ 
    \midrule
    Feedback Signal
      & Implicit/secondary (e.g., success prompts, human ratings)
      & Explicit quantitative rewards from the environment \\
    \midrule
    Non‑Stationarity
      & Less central (fixed weights)
      & Policies co‑evolution increases non‑stationarity, (need for stabilization techniques)\\
    \midrule
    Scalability issues
      & Inference cost, prompt length, loss of coherence
      & Combinatorial explosion of joint action space and state space \\
    \midrule
     Emergent Behavior
      & From pre‑learned language priors (no new behaviors learned via weight update; novel collective phenomena emerge through interaction)
      & Through policy co‑adaptation driven by exploration and rewards\\
    \midrule
     Evaluation Metrics
      & Qualitative (coherence, user satisfaction) and task success
      & Quantitative (cumulative reward, convergence, sample efficiency) \\
    \bottomrule
  \end{tabular}
  \end{small}
\end{table}

\section{An interactionist paradigm to study generative agents' collective behaviors}
\label{sec:roadmap}
\begin{figure}[ht]
    \centering
    \includegraphics[width=1\linewidth]{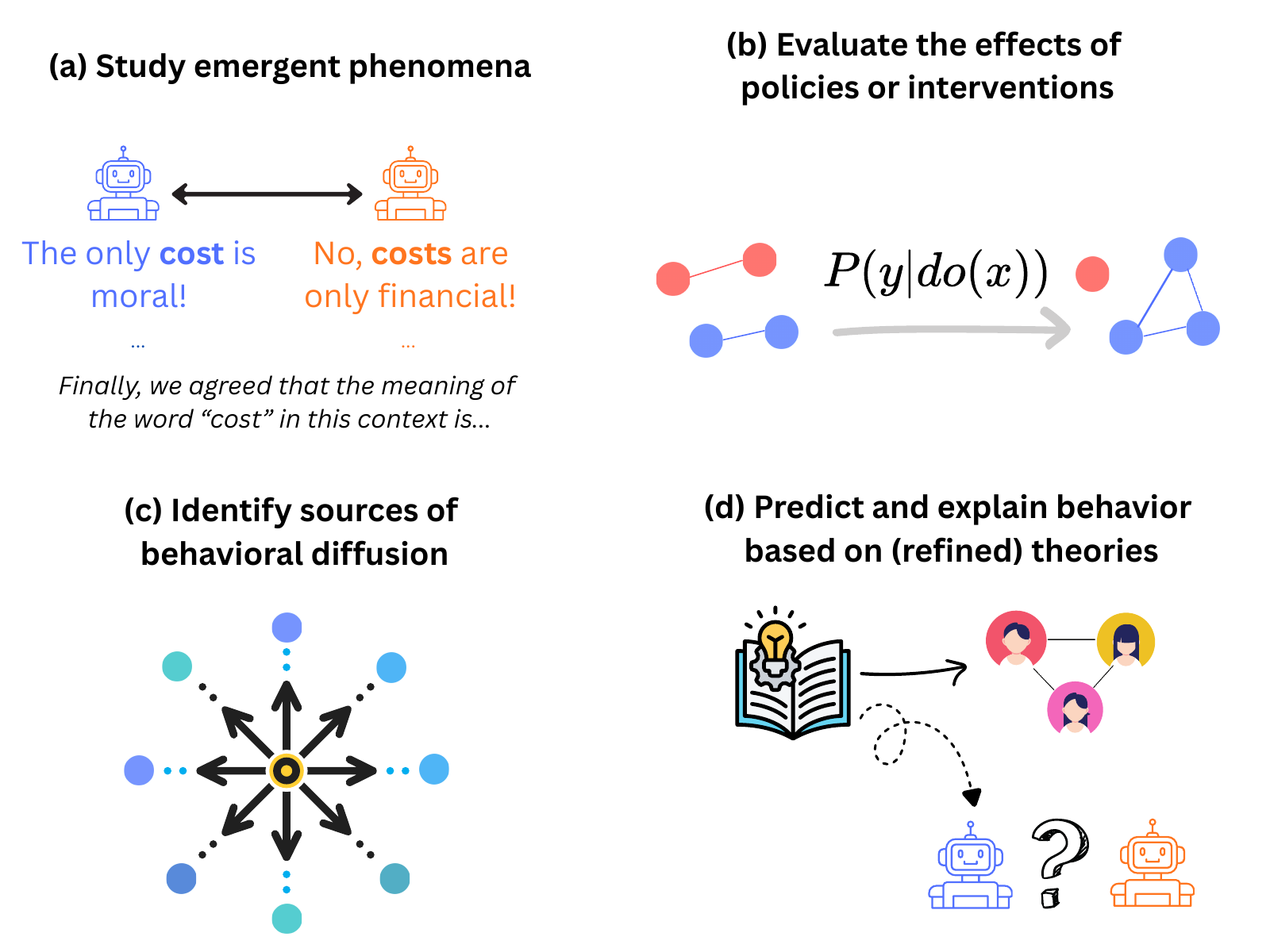}
    \caption{\textbf{Four areas in which our paradigm would be critical}: (a) the study of emergent phenomena, exemplified by the analysis of shared meaning of concepts in debates, through the interactionist lens; (b) the evaluation policies applied to MAS via causal inference; (c) the identification of sources of behavioral diffusion in MAS via information-theoretic insights and causal inference design; (d) the development of theories via empirical scrutiny building on sociological frameworks originally designed for humans.}
    \label{fig:paradigm_use_examples}
\end{figure}

To make progress in the study and deployment of socially interactive generative agents, we propose a paradigm grounded in four foundational dimensions. Each dimension addresses a critical gap in our current understanding of how generative agents behave individually and collectively in dynamic social contexts. Together, they offer a comprehensive framework for analyzing, explaining, and guiding emergent machine behaviors in a rapidly evolving technological landscape, as exemplified in Figure \ref{fig:paradigm_use_examples}. 

\paragraph{Towards an interactionist theory for Gen-AI agents' behavior.} 
Inspired by the literature on the person-situation debate \citep{epstein1985, kenrich1988}, we propose the development of an \textit{interactionist theory} \citep{furr2021} to analyze the determinants of Gen-AI agents' collective behaviors. While it is widely accepted that human actions depend both on individual characteristics (e.g., attitudes, traits, values) and the situations they encounter, social and behavioral sciences have long featured two competing perspectives: (i) the \textit{person-perspective}, which holds that behavior is primarily driven by stable traits and dispositions, and (ii) the \textit{situation-perspective}, which attributes behavior to contextual factors.

This interactionist synthesis draws on deeper sociological foundations. \cite{mead1934mind} distinguished between the ``I'' (the spontaneous, creative aspect of self) and the ``me'' (internalized social attitudes), which map directly onto generative agent architectures. The ``me'' corresponds to pre-trained priors---what Mead termed the ``generalized other,'' the internalized attitudes of the social community acquired through training on human text. The ``I'' corresponds to the contextual, generative response. \cite{cooley1902looking} proposed the concept of the looking-glass self---wherein self-concept emerges through imagined appraisals of others---illuminates in-context learning dynamics. LLM agents continuously ``read'' the social situation and adjust their performed self accordingly. \cite{goffman1959presentation}, through its dramaturgical framework, further clarifies: prompt-based role-playing constitutes front-stage performance, while parametric knowledge forms the backstage repertoire selectively activated by situational demands.
Notably, this synthesis has also been increasingly applied to the study of delinquency and criminal behavior, integrating biological, psychological, economic, and social factors with environmental and institutional interactions \citep{ThornberryInteractionalTheoryDelinquency1987, WrightEffectsSocialTies2001, ZimmermanImpulsivityOffendingNeighborhood2010}.

Turning to Gen-AI agents, they enter any context with extensive pre-trained knowledge, encoded in their parameters through pre-training, SFT, and RLHF. Through ICL, then, such corpus of knowledge is employed, reacting to specific queries and resulting in the emergence of situation-based adaptation of behaviors. As shown in \citet{vonoswald2023transformerslearnincontextgradient}, transformers' forward computations produce behaviors that are similar to those resulting from gradient-based learning steps performed on the examples provided in the context. Crucially, however, interaction currently influences behavior only within the active context; it does not correspond to permanent updates of the model parameters, and hence does not permanently modify the agent’s \emph{parametric} internal priors \citep{dherin2025learningtrainingimplicitdynamics}. In other words, while the model behavior may internally simulate the effect of gradient-style updates, this process is part of the transient computation induced by the current input context, not a lasting change to the agent itself. Once removed from a context and deployed elsewhere, previously emergent behaviors may disappear, especially if contextual data is not transferred--or evolve, depending on the agent’s capacity to retrieve prior contextual information, an area of ongoing research spurred by recent progress in factual memory retrieval \citep{liu2023lostmiddlelanguagemodels}.

Pioneer empirical works already contain indications of persona-situation effects in generative models. At the single-agent level, \citet{bigelow2025beliefdynamicsrevealdual} demonstrate that both direct model modification through steering vectors, and contextual manipulation via prompt engineering, significantly influence model behavior at inference time. The authors argue that these interventions operate by altering the model belief over latent concepts. Beyond individual agents, recent studies on social contexts have adopted a similar view. \citet{Ashery2025} show the emergence of collective bias due to social interactions between generative agents even when the agents themselves had no individual bias. By jointly considering personal and social factors, their work illustrates how interaction dynamics can override individual characteristics. Similarly, \citet{PiaoEmergencehumanlikepolarization2025} highlight the promises of an interactionist framework for studying the complex behaviors of collective AI systems composed of generative agents. The importance of the situation factors in interactive settings is further emphasized by \citet{liu2025spiral}, who demonstrate that self-play in multi-turn, zero-sum games can foster enhanced reasoning abilities, without relying on human-curated data or engineered rewards.

This calls for a perspective that views generative collective behavioral outcomes as shaped by the interplay between internalized priors (the ``person''-perspective) and the interactive, in-context social learning phase (the ``situation''-perspective). In our vision, interactionist theory is a promising framework for understanding the roots of Gen-AI collective behaviors, not as mere emergent side effects, but as adaptive mechanisms that are central in collective, social, and cultural evolution \citep{krafft2021bayesian}. We believe the field would benefit from a rich and systematic literature examining how individual priors, biases, and pre-trained knowledge influence interactive scenarios and shape group outcomes, as well as the symmetric problem of how social interactions among agents can reciprocally influence the priors and traits of individual agents in future contexts. A key open question is how collective behaviors rising through continual interaction might in turn modify or reshape agents’ pre-trained personas, enabling a form of community-level continual learning, layered on top of fixed models, and echoing evolutionary mechanisms.

\paragraph{The need for causality.}
Causality has historically played a limited role in AI research, aside from foundational work arguing that causal reasoning is essential for intelligence \citep{Spirtesprobabilitycausality1991, PearlCausality2009c}. Recently, it has gained renewed attention in areas such as explainability, accountability, and decision-making \citep{JanzingFeaturerelevancequantification2019, ScholkopfCausalRepresentationLearning2021, MadumalExplainableReinforcementLearning2020, Verma2024Counterfactual}. We argue that causality must now become central to the study of interactive generative AI systems. This role extends beyond embedding causal reasoning within individual agents. Causal inference is required to understand emergent phenomena arising from interactions among LLM-based agents—phenomena that cannot be explained by single-agent behavior alone. 

Two classes of causal problems are particularly salient. First, causal methods are needed to identify the mechanisms driving the emergence and diffusion of malicious, deviant, or unintended behaviors in multi-agent systems. This includes attributing responsibility to specific agents, traits, or interaction patterns, and isolating pathways through which beliefs, norms, or values propagate. Such attribution is challenging due to interference, where one agent’s actions affect others, rendering standard causal assumptions invalid without principled designs \citep{BhattacharyaCausalInferenceInterference2020}. Second, causal inference provides the tools required to evaluate interventions and policies \citep{AtheyMachineLearningCausal2015b}. In interactive systems, policies include safety constraints, alignment strategies, or governance rules imposed on agents or interactions. Mapping interventions to downstream collective behavior is essential for assessing both their effectiveness and unintended consequences.

Methodologically, generative AI collectives differ from traditional causal settings in important ways. They offer much higher flexibility to partially mitigate the ``fundamental problem of causal inference'' \citep{HollandStatisticsCausalInference1986b}, enabling \textit{in silico} experimentation, for instance. Researchers can manipulate prompts, replay interactions, and approximate counterfactuals while tracking fine-grained influence across agents. Their dynamic and stochastic nature further motivates approaches such as causal discovery over interaction graphs and causal representation learning over latent semantic states \citep{ScholkopfCausalRepresentationLearning2021, AhujaInterventionalCausalRepresentation2023}.

Beyond scientific understanding, causal analysis is essential for governance and accountability. Without identifying causal pathways, it is difficult to assign responsibility for harmful outcomes or to design effective safeguards \citep{LehmannCausationAILaw2004}. Causal inference has long equipped the social sciences with tools to study behavior and inform policy, including in settings with interference \citep{AngristCredibilityRevolutionEmpirical2010, OgburnCausalDiagramsInterference2014,  AtheyStateAppliedEconometrics2017a, AronowEstimatingaveragecausal2017}. As a field, AI must now take greater responsibility for developing causal frameworks that address both the internal notions of cause and effect within agents and the causal relationships governing their interactions. These considerations motivate the need for benchmarks, evaluation protocols, and experimental designs explicitly tailored to causal questions in multi-agent generative AI.

\paragraph{Information-theoretic insights.}
Information-theoretic measures, such as mutual information, entropy, and information flow, can offer insight into how knowledge is distributed and propagated among generative agents' networks \citep{e24121719}. In particular, in the context of generative agents, mutual information among outputs within a collective can provide a measure to evaluate the presence of shared representations and norms. At the same time, entropy across agents' outputs may reflect the level of innovation and adaptability within the population. Finally, information flow, if measured in terms of transfer entropy, might lead to valuable observations on causal relationships and influence patterns within agents, identifying the emergence of dynamics such as leader/follower or information cascades. With validation against synthetic benchmarks and cross-representation comparisons, these tools might allow us to quantify coordination, redundancy, influence, and innovation across collectives of interacting LLMs, providing a rigorous language to describe the emergence of behaviors.

Undoubtedly, applying information-theoretic measures to LLM collectives presents challenges, for instance, representing agent outputs (tokens, embeddings, or probability distributions), estimating quantities like entropy or mutual information in high-dimensional spaces, and ensuring computational scalability when many agents interact over long conversations. Causal measures such as transfer entropy also require careful handling of temporal structure. Yet, recent advances in variational and embedding-based estimators \citep{belghazi2021minemutualinformationneural, letizia2024mutualinformationestimationfdivergence}, along with scalable strategies \citep{gowri2024approximatingmutualinformationhighdimensional, Kalafut2022.10.15.512388} make these analyses increasingly practical.

\paragraph{AI beyond AI: A sociology of machines.} 
Finally, we argue that AI research must actively engage with neighboring disciplines that offer mature tools for studying emergence, coordination, and deviance in complex systems. 

In particular, we advocate for developing a \textit{sociology of machines}. In the social sciences, AI has largely been treated as a tool for analysis \citep{MolinaMachineLearningSociology2019d, GrimmerMachineLearningSocial2021a}, or more recently as a means to simulate human behavior \citep{FilippasLargeLanguageModels2024, SreedharSimulatingHumanStrategic2024, XieCanlargelanguage2024, AnthisLLMSocialSimulations2025}. We argue that this perspective is no longer sufficient. As AI agents increasingly exhibit autonomy and operate in shared environments \citep{CAIF_1, FloridiAIAgencyIntelligence2025}, they must be studied as social actors in their own right.

This shift builds on early sociological critiques of viewing machines as passive artifacts \citep{WoolgarWhynotSociology1985b} and on Actor–Network Theory, which treated social order as emerging from networks of interacting human and non-human actors \citep{LawActorNetworkTheory1999, LatourReassemblingSocialIntroduction2005}. Early efforts to connect AI and sociological theory followed in the 1990s \citep{Carleynaturesocialagent1994, CarleyArtificialIntelligenceSociology1996a}. Much of this lineage, however, remained largely theoretical or conceptual: we argue that it should now be operationalized to analyze contemporary multi-agent AI systems, building on more recent works that empirically advocated for sociological frameworks involving humans and machines alike \citep{RahwanMachinebehaviour2019, Brinkmann_2023, Tsvetkovanewsociologyhumans2024a}. Notably, however, this recent scholarship mostly emphasizes how machines influence humans. We instead underscore the growing need for perspectives centered on machine-machine interaction \citep{AiroldiMachineHabitusSociology2021, CollinsWhyartificialintelligence2025, CampedelliCriminologyMachines2025}. 

This need is increasingly urgent as LLM-based multi-agent systems are deployed at scale, including in high-stakes domains \citep{CAIF_1, deWittOpenChallengesMultiAgent2025}. Even in the absence of direct human involvement, such systems can exhibit emergent dynamics that diverge from human expectations \citep{CampedelliCriminologyMachines2025}. Crucially, their growing deployment creates an opportunity to move beyond speculative accounts toward systematic empirical study.

A sociology of machines would provide concrete tools for studying how interaction structures, roles, incentives, and learning dynamics among Gen-AI agents give rise to collective outcomes, including coordination, conflict, norm formation, and deviance. Methodologically, a sociology of machines enables testable hypotheses about collective AI behavior and supports empirical strategies such as controlled experimentation, interaction logging, and causal analysis. By treating agent interactions as social processes, researchers can better attribute responsibility for emergent outcomes, detect early signs of harmful dynamics, and evaluate interventions aimed at steering collective behavior. This perspective also helps address the opacity of complex multi-agent systems by situating individual agent actions within broader interactional contexts.

Finally, while human sociological theories offer valuable starting points, they are unlikely to fully explain or predict the dynamics of autonomous AI collectives. Human sociology should therefore serve as a foundation for developing a distinct, empirically grounded sociology of machines—one capable of evolving alongside the rapidly changing architectures and capabilities of generative agents.

\section{Alternative Views}\label{sec:alternative}
Three main alternative views are relevant to our proposed paradigm. We discuss them below. 

\paragraph{LLMs Miss the Multi-Agent Mark.}
\citet{MalfaLargeLanguageModels2025} argue that current LLM-based multi-agent systems (MAS) fail to capture core multi-agent properties and should align more closely with established MAS principles rather than motivate new frameworks. They note that LLMs lack native social behavior 
, attribute many MAS failures to inter-agent misalignment, and advocate for explicit multi-agent pre-training in agent-agnostic environments grounded in structured protocols and formal methods. These critiques highlight real challenges, but our interactionist paradigm addresses complementary questions largely independent of specific architectures or training regimes. Rather than prescribing agent design, it examines how agents with prior representations behave in social contexts, how influence and information propagate, and how collective phenomena emerge. Explicit social pre-training is one design choice, but emergence and alignment dynamics remain relevant across implementations. Our approach thus extends classical MAS theory, retaining its rigor while targeting different aspects of multi-agent behavior.

\paragraph{Gen-AI models show uniform collective behavior.}
Shared training data and fine-tuning methods, being optimized to satisfy similar human preferences and safety constraints, make individual models converge to the same dominant behavior, rather than exploring different possibilities \citep{jiang2025artificialhivemindopenendedhomogeneity}. While this might be plausible for models acting in isolation, it is substantially less reasonable in interactive MAS. Even agents with identical pre-training can produce diverse collective outcomes, shaped by initial conditions, interaction structures, and contextual factors. In practice, agents vary in fine-tuning, prompting, retrieval context, and interaction history, and multi-agent populations may include heterogeneous model architectures altogether. Our framework captures how subtle differences in the ``person'' component (pre-trained priors) interact with the ``situation'' component (social context) to produce emergent phenomena. If convergence to uniform behaviors were unavoidable, we would not observe the range of social conventions, polarization, and coordination patterns documented in recent Gen-AI MAS studies \citep{PiaoEmergencehumanlikepolarization2025,Ashery2025, FlintGroupsizeeffects2025}.

\paragraph{Gen-AI agents still fail at simple tasks.}
Existing LLMs still struggle with relatively simple tasks \citep{WilliamsEasyProblemsThat2024, MalekFrontierLLMsStill2025, XuHallucinationInevitableInnate2025}, and recent work demonstrated that the performance of MAS with LLM-powered agents is outperformed by single agents across a variety of tasks \citep{PanWhyMultiagentSystems2025a}. Yet, these failures underscore rather than diminish the importance of rigorous analysis. Understanding why multi-agent systems fail requires causal attribution, measurement of information flow breakdowns, and identification of coordination failures—exactly what the interactionist paradigm provides. Societal deployment of these systems is already outpacing theory \citep{CAIF_1, deWittOpenChallengesMultiAgent2025}, so waiting for more capable agents before developing frameworks would be irresponsible. Current failures are valuable data for theory development that will remain relevant as capabilities grow. Our paradigm focuses on the mechanisms driving agent collectives, independent of individual agent sophistication.

\section{Conclusions}
\label{sec:concl}
In this perspective, we have argued that understanding the collective behavior of Gen-AI agents is an urgent challenge with far-reaching societal implications. Generative agents differ significantly from traditional agents due to their initialization with vast pre-trained knowledge, implicit social priors, and capacity for in-context adaptation. These features give rise to complex, emergent behaviors when such agents interact, behaviors that current theoretical tools, developed for the study of interaction in MARL, are ill-equipped to fully explain or manage. To support this endeavor, we have proposed the need for a new framework, grounded in transdisciplinary dialogue, drawing on insights from cognitive science, social and cultural learning theory, and machine learning. We envision concrete research directions:
\begin{enumerate}
  \item \textbf{Interactionist benchmarks}: Evaluation protocols designed to isolate contributions of ``situation'' (prompt context, interaction history) versus ``person'' (pre-trained priors, model scale, alignment).
  \item \textbf{Causal identification strategies}: Experimental and quasi-experimental designs for identifying causal pathways in networked LLM systems, building on interference-aware methods \citep{AronowEstimatingaveragecausal2017, BhattacharyaCausalInferenceInterference2020}.
  \item \textbf{Information-theoretic measures}: Operationalizations of influence, consensus, and innovation flow within agent collectives. \item \textbf{Empirical sociology of machine societies}: Using LLM collectives as model organisms for studying social processes under controlled conditions.
\end{enumerate}
Such a framework can help us better anticipate and steer the emergent dynamics of generative agents, whether the goal is to foster beneficial cooperation, mitigate risk, or ensure alignment with human values.

\section*{Acknowledgments}
This work was partially supported by the following projects: Horizon Europe Programme, grants \#10112- 0237-ELIAS and \#101120763-TANGO. Funded by the European Union. Views and opinions expressed are however those of the author(s) only and do not necessarily reflect those of the European Union or the European Health and Digital Executive Agency (HaDEA). Neither the European Union nor the granting authority can be held responsible for them.

This work was also partly supported by Ministero delle Imprese e del Made in Italy (IPCEI Cloud DM 27 giugno 2022 – IPCEI-CL-0000007) and European Union (Next Generation EU).

\bibliography{mlfeo}
\bibliographystyle{apalike}

\end{document}